\RequirePackage{amsmath}
\documentclass[runningheads]{llncs}
\usepackage[T1]{fontenc}
\usepackage[pdftex]{graphicx}
\graphicspath{{./images/}}
\DeclareGraphicsExtensions{.pdf,.jpg,.png}

\usepackage{booktabs}
\usepackage{multirow}

\usepackage{amsmath}
\usepackage{amssymb}

\usepackage{algorithmic}
\usepackage{algorithm}

\usepackage{array}

\usepackage[caption=false,font=footnotesize,captionskip=5pt]{subfig}

\usepackage{stfloats}

\usepackage{url}
\usepackage{hyperref}

\begin{document}

\title{DualLQR: Efficient Grasping of Oscillating Apples using Task Parameterized Learning from Demonstration}
\titlerunning{DualLQR: Grasping Oscillating Apples}

\author{Robert~van~de~Ven\inst{1} \and
        Ard~Nieuwenhuizen\inst{2} \and
        Eldert~J.~van~Henten\inst{1} \and
        and~Gert~Kootstra\inst{1}}

\authorrunning{R. van de Ven et al.}

\institute{
    Agricultural Biosystems Engineering, Wageningen University and Research, 6708PB, Wageningen, The Netherlands \and
    Agrosystems Research, Wageningen University \& Research, 6708 PB Wageningen, the Netherlands\\
    \email{robert.vandeven@wur.nl}
}

\maketitle

\begin{abstract}
Learning from Demonstration offers great potential for robots to learn to perform agricultural tasks, specifically selective harvesting. One of the challenges is that the target fruit can be oscillating while approaching. Grasping oscillating targets has two requirements: 1)~close tracking of the target during the final approach for damage-free grasping, and 2)~the complete path should be as short as possible for improved efficiency. We propose a new method called DualLQR. In this method, we use a finite horizon Linear Quadratic Regulator~(LQR) on a moving target, without the need of refitting the LQR. To make this possible, we use a dual LQR set-up, with an LQR running in two separate reference frames.
Through extensive simulation testing, it was found that the state-of-art method barely meets the required final accuracy without oscillations and drops below the required accuracy with an oscillating target. DualLQR, on the other hand, was found to be able to meet the required final accuracy even with high oscillations, while travelling the least distance.
Further testing on a real-world apple grasping task showed that DualLQR was able to successfully grasp oscillating apples, with a success rate of 99\%.

\keywords{Learning from Demonstration  \and Oscillating \and Apple grasping}
\end{abstract}

\section{Introduction}
Learning from Demonstration~(LfD) offers great potential for robots to learn to perform tasks through human demonstrations, removing the need for programming complex motions into robots, which is a time-expensive task in variable environments~\cite{RN101}. 
LfD has been implemented for various tasks, such as in-home assistance~\cite{RN169}, in agricultural~\cite{RN179,RN142,RN166}, and industrial tasks~\cite{RN186}. However, many domain-specific challenges remain unaddressed.

Within agricultural tasks, LfD is valuable for selective harvesting. Selective harvesting is challenging to automate because of variation and specific detachment motions required for safe harvesting of different types of fruits~\cite{RN153}. In selective harvesting, manipulators are used in three phases: 1)~approach to target fruit, 2)~detachment of fruit from the plant, and 3)~placement of fruit in a storage container~\cite{RN190}. In the approach phase, the manipulator must approach the fruit safely, without damaging the plant or the fruit. 
One of the challenges during the approach phase is a moving target fruit~\cite{RN174}. This moving can occur because of hitting a compliant branch during the harvest operation, because of removing other fruits from the same branch, or because of wind. Due to the compliance in a branch, this can result in long-lasting oscillations of the fruit. To still be able to grasp the fruit, the robot needs to adapt to the oscillation. 
To deal with the oscillation of a target, visual servo control has been used, but has trouble dealing with unforeseen interactions~\cite{RN193}. 
This challenge of oscillating targets has not yet been addressed and solved in harvesting robotics~\cite{RN174}, let alone for LfD in selective harvesting.

In LfD, tracking moving targets has been studied before. Research has focused on grasping moving objects from conveyors~\cite{RN186,RN187}, catching or hitting flying objects~\cite{RN188,RN183}, or grasping a moving target during the task execution~\cite{RN176,RN185,RN175}. However, these studies did not focus on tracking the target only when close to grasping and the target's movement was not oscillatory. 
For selective harvesting, oscillations are the most common movements and tracking should only occur when close to the target, to reduce the length of the path and be able to speed up the task execution. Therefore, an LfD algorithm for selective harvesting has two requirements: 1)~close tracking of the target during the final approach for damage-free grasping, and 2)~the complete path should be as short as possible for improved efficiency. 
The combination of these two requirements adds a requirement for information on when tracking is important.

To combine these requirements, the task can be learned relative to the pose of the moving target for close tracking during the final approach, and relative to the initial pose of the end-effector for reduced tracking behaviour when further away. These poses are then used as reference frames to transform the demonstrations to. By transforming the demonstrations to each reference frame, the variation in the demonstrations in each reference frame can be used as information on when tracking is important. 
Task Parameterized Gaussian Mixture Models~(TP-GMM) combine these requirements~\cite{RN175}. 
Task Parameterization~(TP) allows a task to be learned using multiple reference frames, transforming the demonstrations to each reference frame. In each reference frame, a GMM can be used to learn the variability of the demonstrations. By using multiple Gaussians, the variability can be learned in higher detail, e.g. variation over time and space. To transform the TP-GMM to trajectories that can be used by a robot, a regression method needs to be used, e.g. Gaussian Mixture Regression~(GMR) and/or Linear Quadratic Regulator~(LQR)~\cite{RN26}. 

Calinon et al.~\cite{RN175} proposed a TP-GMM and GMR, coupled to an infinite horizon LQR~(InfLQR)  to deal with targets moving during execution. At each timestep, the method performs three steps: 1)~combine the GMMs in each reference frame given the new poses of the reference frames, 2)~perform GMR on the combined GMM to extract the end-effector goal and covariance matrix, and 3)~calculate control commands based on the output of the GMR using the InfLQR. By using an InfLQR, the method can calculate control commands without calculating the remaining trajectory at each timestep. 
The authors showed that the algorithm could execute the task with a moving target. However, no analysis of the performance was performed. 
InfLQR was used to limit computation time. This was important since the moving target would result in changes in the combined GMM, which would mean that a finite horizon LQR would need to be refitted on the complete remaining part of the trajectory. However, a finite horizon LQR is preferred since it can consider the varying precision in the remaining part of the task~\cite{RN175}.

We propose a new method called DualLQR\footnote{Our code is available at \url{https://github.com/WUR-ABE/DualLQR}}. In this method, we use a finite horizon LQR on a moving target, without the need of refitting the LQR, thereby reducing the computations in the real time loop. 
To enable this, we use a dual LQR set-up, with an LQR running in two reference frames. 
During execution, each LQR is used to calculate the desired control in that reference frame. 
These controls are combined using a weighted average based on the precision matrix of the GMR in that reference frame.
For the apple harvesting task, one reference frame is the stationary initial pose of the end-effector and the other reference frame is the oscillating fruit.
The effect of tight tracking during the final approach is achieved in two ways. 
First, each LQR will track tighter if the covariance is smaller, which will happen when the target pose is closer to the origin of the reference frame. 
Second, this effect is enhanced by using the weighted average of precision matrices. Therefore, the LQR in the reference frame with higher required precision will affect the combined control output more.

To evaluate the proposed method, we performed comprehensive simulation experiments testing the proposed method DualLQR and two existing methods, InfLQR~\cite{RN175} and using the LQR using a single reference frame. In addition, we tested the proposed method on a real robot on a simplified apple grasping task.

\section{Background}
In this section, we explain the existing methods on which DualLQR builds. First, we describe the Task-Parameterization to describe the task in reference frames. Next, we describe the InfLQR method~\cite{RN175}. Finally, we describe the LQR using a single reference frame. 

\subsection{Task-Parameterization}
In this section, we explain the Task-Parameterization (TP). Initially, demonstrations were recorded, resulting in trajectories \(X_{g} = [x_{1}, \dots, x_{T}] \in \mathbb{R}^{D\times T}\) in a global frame \(g\) and consisted of \(T\) timesteps. 
In our case, \(D=6\) consisting of a 3-dimensional Cartesian position and 3-dimensional Tait-Bryan angles for each timestep. 
These demonstrations were converted into reference frame \(j\) using transformation matrices, resulting in trajectories \(X_j\). In the reference frame \(j\), the task can be encoded as \(\xi_{j} = [\tau, X_{j}]^{\top}\), with \(\tau = [1, \dots, T]\) and \(\xi_{j} \in \mathbb{R}^{\left(1 + D \right) \times T}\). 

The task  can be learned in all reference frames \(J\) using a single GMM by combining the task encodings of each reference frame into the encoding \(\xi = [\tau, X_{1}, \dots, X_{J}]^{\top}\), with \(\xi \in \mathbb{R}^{\left(1 + J \cdot D \right) \times T}\)

In order to combine reference frames, the GMMs need to be transformed into the same reference frame. To transform a GMM from reference frame \(j\) into the global frame \(g\), a linear transformation is performed. Each reference frames can be defined as an origin vector \(b \in \mathbb{R}^{D+1}\) and a transformation matrix \(A \in \mathbb{R}^{\left( D+1 \right) \times \left( D+1 \right)}\). For reference frame \(j\), this is defined as \cite{RN175}:
\begin{equation}\label{eq:tp-transform}
    \xi_{j} = \begin{bmatrix}
        \tau \\ X^p_j \\ X^r_j
    \end{bmatrix}
    \text{, }
    b_{j} = \begin{bmatrix}
        0 \\ x^p_j \\ x^r_j
    \end{bmatrix}
    \text{, }
    A_{j} = \begin{bmatrix}
        I & 0 & 0 \\ 0 & R_j & 0 \\ 0 & 0 & I
    \end{bmatrix}
\end{equation}
where \(\tau\) is the time dimension, \(X^p_j\) is the trajectory of the 3-dimensional Cartesian position and \(X^r_j\) is the trajectory of the 3-dimensional orientation represented using Tait-Bryan angles in frame \(j\). \(x^p_j\) indicates the 3-dimensional position and \(x^r_j\) and \(R_j\) indicate the 3-dimensional orientation, using the Tait-Bryan angles and rotation matrix of frame \(j\). \(0\) are zeros vectors/matrices of the corresponding size and \(I\) identity matrices of the corresponding size. 

\subsection{InfLQR}
In this section, we describe how the infinite horizon LQR is initialized and fitted before execution, and how the method is used during execution. 

\subsubsection{Before execution}
In InfLQR, the GMM is fitted before execution. 
A GMM combines \(K\) Gaussian distributions in order to represent the demonstrated trajectories. The component \(C_{k}\) is a Gaussian distribution \(\mathcal{N}(\mu_{k}, \Sigma_{k})\), using the task encoding \(\xi\). 
These Gaussian components were fitted to the data using an Expectation-Maximization (EM) algorithm \cite{RN25}. 

\subsubsection{During execution}
At each timestep \(t\) during the execution, several steps need to be performed to get a GMM in the global frame. 
As the task was encoded using \(\xi\), the GMM containing multiple references frames needs to be split into separate GMMs containing a single reference frames, with a task encoding of \(\xi_{j}\) and \(K\) components consisting of \(\mathcal{N}(\mu_{k,j}, \Sigma_{k,j})\). 

As the task is still encoded in each reference frame, the next step is to obtain the global encoding given the current timestep \(t\) and using the current location of the reference frames \(A_{t,j}\) and \(b_{t,j}\), defined as \(\xi_t = [\tau, X_{g}]\). This is calculated using the product of the linearly transformed GMM from each reference frame \(J\), using the following equations \cite{RN175}:

\begin{equation}\label{eq:pog-sigma}
    \Sigma_{k,t} = \left( \sum_{j=1}^{J} \left(A_{t,j} \Sigma_{k,j} A_{t,j}^{\top} \right)^{-1} \right)^{-1}
\end{equation}
\begin{equation}\label{eq:pog-mu}
    \mu_{k,t} = \Sigma_{k,t} \left( \sum_{j=1}^{J} \left( A_{t,j} \Sigma_{k,j} A_{t,j}^{\top} \right)^{-1} \left( A_{t,j} \mu_{k,j} + b_{t,j} \right) \right)
\end{equation}
where \(\mu_{k,j}\) and  \(\Sigma_{k,j}\) are first transformed into the global reference using \(A_{t,j}\) and \(b_{t,j}\), after which the reference frames are added together based on the required precision in each frame. 
This results in a combined GMM for timestep \(t\) with task encoding of \(\xi_{t}\) and \(K\) components consisting of \(\mathcal{N}(\mu_{k,t}, \Sigma_{k,t})\).

The next step is to calculate the target mean and covariance given the combined GMM. For this step, Gaussian Mixture Regression (GMR) is used. GMR uses the components of the GMM to determine the mean and covariance of the output dimensions \(\mathcal{O}\) given a value in the input dimension \(\mathcal{I}\). In this case, the input dimension corresponds to the time dimensions in the task encoding and the output dimension corresponds to the pose in the task encoding. The input and output dimensions are also applied to the components of the GMM, resulting in the following encoding, mean and covariance:
\begin{equation}
    \xi_t = \begin{bmatrix} \xi_{t}^{\mathcal{I}} \\ \xi_{t}^{\mathcal{O}} \end{bmatrix} \text{, }
    \mu_{k,t} = \begin{bmatrix} \mu_{k,t}^{\mathcal{I}} \\ \mu_{k,t}^{\mathcal{O}} \end{bmatrix} \text{, }
    \Sigma_{k,t} = \begin{bmatrix} \Sigma_{k,t}^{\mathcal{I}} & \Sigma_{k,t}^{\mathcal{IO}} \\ \Sigma_{k,t}^{\mathcal{OI}} & \Sigma_{k,t}^{\mathcal{O}} \end{bmatrix}
\end{equation}

At timestep \(t\), \(\xi_{t}^{\mathcal{I}} = t\), and can be used to calculate the target mean and covariance of the pose, using the following equations \cite{RN175}:
\begin{equation}\label{eq:gmr-mu}
    \hat{\mu}_{t}^{\mathcal{O}} = \sum_{k=1}^{K} h_{k}\left(\xi_{t}^\mathcal{I}\right)
    \left( \mu_{k,t}^{\mathcal{O}} + \Sigma_{k,t}^{\mathcal{OI}} {\Sigma_{k,t}^{\mathcal{I}}}^{-1} \left( \xi_{t}^\mathcal{I} - \mu_{k,t}^{\mathcal{I}} \right) \right)
\end{equation}
\begin{equation}\label{eq:gmr-sigma}
    \hat{\Sigma}_{t}^{\mathcal{O}}=\sum_{k=1}^{K} h_{k}\left({\xi_{t}^\mathcal{I}}\right)
    \left(\Sigma_{k,t}^{\mathcal{O}} - \Sigma_{k,t}^{\mathcal{OI}} {\Sigma_{k,t}^{\mathcal{I}}}^{-1} \Sigma_{k,t}^{\mathcal{IO}} \right)
\end{equation}
which depend on the activation function \(h_{k}\):
\begin{equation}\label{eq:gmr-activation}
    h_{k}\left( \xi_{t}^\mathcal{I} \right) = \frac{\pi_{k} \mathcal{N} \left( \xi_{t}^\mathcal{I} | \mu_{k,t}^\mathcal{I}, \Sigma_{k,t}^\mathcal{I} \right)}{\sum_{k=1}^{K} \pi_{k} \mathcal{N} \left( \xi_{t}^\mathcal{I} | \mu_{k,t}^\mathcal{I}, \Sigma_{k,t}^\mathcal{I} \right)}
\end{equation}
which indicates the importance of each component \(k\) of the combined GMM given the input \(\xi_{t}^\mathcal{I}\).

Next, the infinite horizon LQR is used to determine the required control outputs to achieve the target mean. The system in the LQR is defined as \cite{RN26}:
\begin{equation}\label{eq:lqr-system}
    x_{t+1} = A x_t + B u_t
\end{equation}
In our case, \(A = I\) and \(B = I \cdot \Delta t\), with \(\Delta t\) the size of the timestep. 
The infinite horizon LQR balances the control outputs \(u_t\) using the matrices \(R\) for the cost and \(Q_t\) for the required precision \cite{RN25}:

\begin{equation}\label{eq:lqr-costs}
    R = I \cdot 10^{\rho} \text{, } Q_t = \left(\hat{\Sigma}_{t}^{\mathcal{O}}\right)^{-1}
\end{equation}
where \(R \in \mathbb{R}^{D\times D}\) is constant and can be tuned using control cost \(\rho\) and \(Q_t\) is the precision matrix of the GMM. The control \(u_t\) is then determined using the following equation \cite{RN175}:
\begin{equation}\label{eq:lqr-u-inf}
    u_t = K_t^{\mathcal{P}} \left( \hat{\mu}_{t}^{\mathcal{O}} - x_t \right)
\end{equation}
where \(\hat{\mu}_{t}^{\mathcal{O}}\) is estimated by the GMR and \(x_t\) indicates the state of the robot. The feedback gain \(K_t^{\mathcal{P}}\) is calculated using the following equation \cite{RN26}:
\begin{equation}\label{eq:lqr-gain-inf}
    K_t^{\mathcal{P}} = \left( R + B^{\top} P_t B \right)^{-1} B^{\top} P_t A
\end{equation}
which depends on the outcome of the Riccati equation:
\begin{equation}\label{eq:lqr-ric-inf}
    P_t = Q_t - A^{\top} \left( Q_t B \left( B^{\top} Q_t B + R \right)^{-1} B^{\top} Q_t - Q_t \right) A
\end{equation}
which for the infinite horizon LQR only considers the required precision at time \(t\). 
With these equations, the feedback gain \(K_t^{\mathcal{P}}\) will be time-variable and perform tighter tracking of target mean \(\hat{\mu}_{t}^{\mathcal{O}}\) based on the desired precision \(Q_t\). 

\subsection{Single LQR}
Alternatively to combining the reference frames, the task could be performed in a single reference frame using a finite horizon LQR coupled to the moving target. This finite horizon LQR can then be fitted before execution. 

\subsubsection{Before execution}\label{sec:single-lqr-before}
With a single LQR, the task is encoded as \(\xi_{j,t} = [t, x_{j,t}^{\top}]^{\top}\). The GMM was fitted using EM. Since the task is consisted in the frame of the moving target, the GMR can be performed before execution and a finite horizon LQR can be fitted to this trajectory. 

For the GMR, Equations~\ref{eq:gmr-mu} and \ref{eq:gmr-sigma} are used to calculate the desired trajectory \(\hat{\mu}^{\mathcal{O}} = [\hat{\mu}_{0}^{\mathcal{O}} \cdots \hat{\mu}_{T}^{\mathcal{O}} ]\), with variation  \(\hat{\Sigma}^{\mathcal{O}} = [\hat{\Sigma}_{0}^{\mathcal{O}} \cdots \hat{\Sigma}_{T}^{\mathcal{O}} ]\). 

The LQR is then fitted to this trajectory to determine the optimal control outputs to achieve the final target. 
The finite horizon LQR uses a set of recursive functions to determine the feedforward gain \(K_t^{\mathcal{V}}\) and feedback gain \(K_t^{\mathcal{P}}\) \cite{RN26}:
\begin{equation}\label{eq:lqr-Kv}
    K_t^{\mathcal{V}} = \left(R+B^{\top} S_{t+1} B\right)^{-1}B^{\top}
\end{equation}
\begin{equation}\label{eq:lqr-K}
   K_t^{\mathcal{P}} = \left(R + B^{\top} S_{t+1} B \right)^{-1} B^{\top} S_{t+1} A
\end{equation}
\begin{equation}\label{eq:lqr-S}
    S_{t} = A^{\top} v_{t+1} \left(A - B K_t^{\mathcal{P}} \right) + Q_t
\end{equation}
\begin{equation}\label{eq:lqr-v}
    v_{t} = \left( A - B K_{t}^{\mathcal{P}} \right)^{\top} v_{t+1} + Q_t \hat{\mu}_{t}^{\mathcal{O}}
\end{equation}
where \(A\) and \(B\) describe the system as defined in Equation~\ref{eq:lqr-system}, \(R\) is the control cost and \(Q\) is the required precision, both defined in Equation~\ref{eq:lqr-costs}. \(S_{t}\) and \(v_{t}\) are the Riccati and output function respectively, which are used to recursively calculate the gains from the final conditions \(S_T = Q_T\) and \(v_T = Q_T\hat{\mu}_{T}^{\mathcal{O}}\). 

With these equations, the final conditions \(S_T\) and \(v_T\) are used to determine intermediate conditions and gains \(K_t^{\mathcal{V}}\) and \(K_t^{\mathcal{P}}\). These are used by the LQR during the execution to determine control outputs which achieve both the current conditions, but also take the final condition into account. 

\subsubsection{During execution}
At each timestep, the control output can be directly determined from the gains and the output function, calculated in Equations~\ref{eq:lqr-Kv},  \ref{eq:lqr-K}, and \ref{eq:lqr-v}. In addition, the current state \(x_t\) is used. This is the affine optimal control equation \cite{RN26}:
\begin{equation}\label{eq:lqr-u}
    u_{i}=-K_t^{\mathcal{P}} x_{t} + K_t^{\mathcal{V}} v_{t+1}
\end{equation}
where the feedback term \(K_t^{\mathcal{P}}\) allows the control to be determined in terms of the current state \(x_{t}\) and the feedforward term \(K_t^{\mathcal{V}}\) anticipates the desired reference signal.

\section{Proposed method} \label{sec:prop-meth}
When the task is only performed in the frame of the moving target, the number of performed equations during execution is reduced a lot. However, because the task is only performed relative to the target, the controller will always attempt to track the target, which can only be minimized by increasing the cost for tracking. On the other hand, the infinite horizon LQR can take both reference frames into account and thereby reduce the initial amount of tracking, but requires several complex computations for this. 

Therefore, we propose DualLQR, which maintains both reference frames, while only adding minimal additional computation costs compare to using a single reference frame. 

\begin{figure}[t]
    \centering
    \subfloat[Steps before execution. In each step, light grey lines show the demonstrations. Each step was performed separately in the start and end frames. The start frame is indicated in blue and the end frame is indicated in orange. \textbf{(1)} shows each reference frame's GMM, where dots indicate the means and ellipses the covariances of the Gaussian components. \textbf{(2)} shows each reference frame's GMR, where the line indicates the mean of the path and ellipses show the covariance along the path. \textbf{(3)} shows each reference frame's LQR. The \(K_t^{\mathcal{P}}\) and \(K_t^{\mathcal{V}}\) indicate the gains, which are tuned using the GMR and an additional control cost \(\rho\).]{
        \includegraphics[width=0.47\columnwidth]{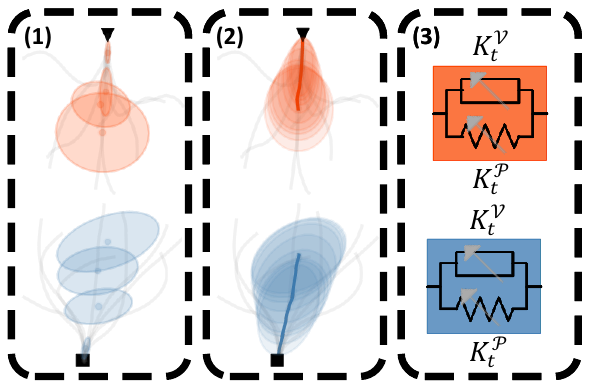}
        \label{fig:flowchart-pre}
    }
    \hfill
    \subfloat[Processes in the control loop during execution visualized in 2D. \textbf{(1)} indicates the task in the world frame, shown in black. The blue frame indicates the start frame and the orange frame indicates the end frame, which oscillates, as the black arrow indicates. The green vector is the control output provided by DualLQR. \textbf{(2)} indicates the task in each reference frame \(j\), transformed using \({}_j^g T_{t}\) and with the covariance matrix \(\hat{\Sigma}_{j,t}^{\mathcal{O}}\) from the GMR. Because of this transformation, the oscillation of the end-frame results in an oscillation of the end-effector in the end-frame.  \textbf{(3)} indicates each LQR, using the end-effector pose \(x_{j,t}\) as input. \textbf{(4)} indicates the balancing between the two LQRs, using \(u_{j,t}\) and \(\hat{\Sigma}_{j,t}^{\mathcal{O}}\), resulting in control output \(u_{t}\).]{
        \includegraphics[width=0.47\columnwidth]{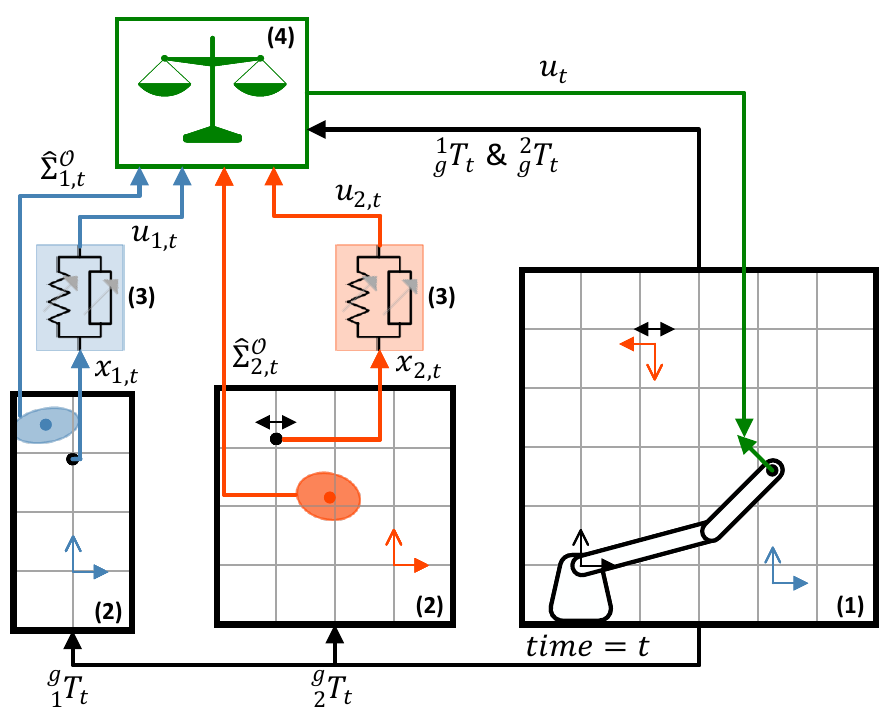}
        \label{fig:flowchart-rt}
    }
    \caption{Flowcharts of the pre-process and real-time processes of DualLQR}
    \label{fig:flowchart}
\end{figure}

\subsubsection{Before execution}\label{sec:before-ex}
For DualLQR, the GMM encoded the task as \(\xi_t = [t, x_{1,t}^{\top}, x_{2,t}^{\top}]^{\top}\). The GMM is fitted using EM. 
Next, a separate GMM is created for each reference frame by splitting along the task dimensions, resulting \(\xi_{1,t} = [t, x_{1,t}^{\top}]^{\top}\) and \(\xi_{2,t} = [t, x_{2,t}^{\top}]^{\top}\). Next, the GMR and LQR are fitted in each reference frame, as described in section \ref{sec:single-lqr-before}. 
The steps performed before execution are shown in Fig.~\ref{fig:flowchart-pre}. 

\subsubsection{During execution}\label{sec:during-ex}
At each timestep and for each reference frame, the control output can be directly determined using Equation~\ref{eq:lqr-u}. 
The outputs of each reference frame are transformed to be in the global reference frame using \(A_i\) from Equation~\ref{eq:tp-transform}. 
In order to combine the control outputs, the weighted average was calculated using the covariance matrix \(\Sigma_{i,t}\) from each GMR \(i\). 
Since we aimed to prioritise the reference frame with the least variance, we used the inverse covariance matrix, i.e. the precision matrix. This resulted in the following equation:
\begin{equation}\label{eq:LQR_weighing}
    u_{t} = \frac{ \sum_{i=1}^{I} \text{diag} \left( \left(\hat{\Sigma}_{i,t}^{\mathcal{O}}\right)^{-1} \right) u_{i,t} }{ \sum_{i=1}^{I} \text{diag} \left( \left(\hat{\Sigma}_{i,t}^{\mathcal{O}}\right)^{-1} \right)}
\end{equation}
with \(u_{i,t}\) being the vector containing controls of LQR \(i\) at timestep \(t\) and \(\Sigma_{i,t}^{\mathcal{O}}\) being the covariance matrix of GMR \(i\) at timestep \(t\). In our case, \(i\) can be \(1\) and \(2\), indicating the start and end reference frames. Lastly, \(u_{t}\) is the vector containing the combined and weighted controls. The covariance matrix provided by the GMR is non-sparse, which is used for the fitting of the LQR. For combining the outputs, only the weighted average needs to be determined. As the control outputs are determined for the correct output dimension, the weighted average should not affect this. Therefore, the covariance matrices were changed to diagonal matrices by setting all non-diagonal values to zero. This follows the convention of matrix \(B\) of Equation~\ref{eq:lqr-system}. 
The steps performed during execution are shown in Fig.~\ref{fig:flowchart-rt}. 

\section{Experimental set-up} \label{sec:experiments}
We performed two experiments. First, we tested both InfLQR~\cite{RN175} and DualLQR in a simple simulation environment with different oscillations and controller settings. Next, we tested DualLQR in a real-world apple grasping task. 

\begin{figure}[t]
    \centering
    \subfloat[Top view]{
        \includegraphics[width=0.4\columnwidth]{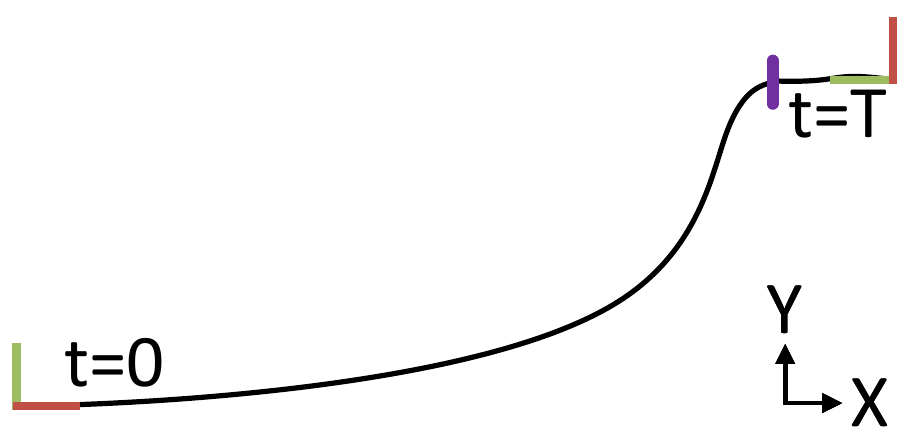}
        \label{fig:task-example-XY}
    }
    \subfloat[Side view]{
        \includegraphics[width=0.4\columnwidth]{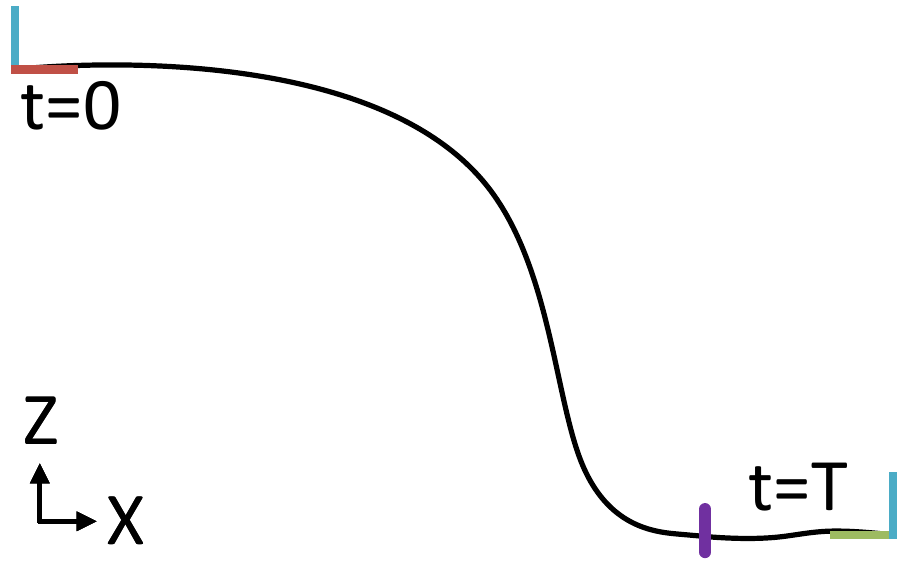}
        \label{fig:task-example-XZ}
    }
    \addtolength{\abovecaptionskip}{0.05in}
    \caption{Typical example of a demonstration. The wireframes indicate the start and end-frame. The purple line indicates the start of the final approach.}
    \label{fig:task-example}
\end{figure}

For both experiments, we used a model trained on 40 demonstrations. In these demonstrations, the target was a 10 cm cube, which was static. A static target was used both to simplify the process of collecting demonstrations, and because the variable conditions in the orchard can result in static targets in all demonstrations. Fig.~\ref{fig:task-example} shows the path of a typical demonstration, going from \(t=0\) to \(t=T\). The same demonstration dataset was used as in Van de Ven et al.~\cite{RN203}.

\subsection{Set-up of simulation experiment} \label{sec:sim-exp}
In the simulation experiment, we tested the infinite horizon LQR~(InfLQR)~\cite{RN175} and our proposed DualLQR. 
The simulation was created using ROS2, using a simulated UR5e manipulator. The moving target was created by defining a target pose with sinusoidal movement in one dimension. 
In this experiment, we adjusted the control cost \(\rho\), the axis along which the oscillation took place and the amplitude of the oscillation.
The control cost \(\rho\) allowed optimization between the closeness of tracking and the total distance travelled by the manipulator, as indicated in Equation~\ref{eq:lqr-costs}. 
The control cost \(\rho\) was adjusted between \(-3.0\) and \(3.0\) in steps of \(0.3\).
Through using different movement axes and amplitudes, we investigated the robustness and which oscillations were the most challenging for the methods. 
The oscillation was performed along x, y, z, roll, pitch, and yaw in the object coordinate frame. For grasping apples, the orientation is also important. While the apple itself is roughly spherical, it is attached at the peduncle, and requires a rotation around this peduncle for correct detachment. While any orientation of the end-effector could still be used to grasp the fruit, the detachment motion has the least influence on the workspace if it can be performed with the final joint of the manipulator. Because of this, the orientation of the fruit needs to be tracked.  
For the results, we grouped these into position or orientation oscillations.
The oscillation amplitude was set at four levels, specifically no oscillation, low oscillation of 0.05 m or 0.15 rad, medium oscillation of 0.10 m or 0.30 rad, and high oscillation of 0.15 m or 0.45 rad.

In each test, we performed eleven repetitions, placing the goal in a new pose each time. We used one central pose at \([0.22, 0.27, -0.26, 0.00, 0.00, 1.46]\). The remaining ten poses changed one dimension, using the following values: X: \(\pm0.1\), Y: \(\pm0.2\), Z: \(\pm0.05\), roll: \(\pm0.2\), yaw: \(\pm0.2\). These poses were based on the variation in the demonstration dataset, where there was minimal variation along the pitch axis. 

\begin{figure}[t]
    \centering
    \subfloat[Top view]{
        \includegraphics[width=0.3\columnwidth]{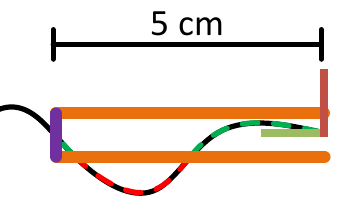}
        \label{fig:final-approach-example-XY}
    }
    \subfloat[Side view]{
        \includegraphics[width=0.3\columnwidth]{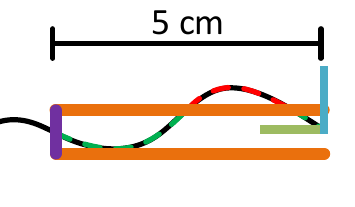}
        \label{fig:final-approach-example-XZ}
    }
    \addtolength{\abovecaptionskip}{0.05in}
    \caption{Example of final approach. The wireframes indicate the end-frame. The black line indicates the trajectory. The purple line at 5 cm from the target indicates the start of the final approach. The orange lines indicate the boundaries. The trajectory is dashed green if it lies within the boundaries. Else, it is dashed red.}
    \label{fig:final-approach-example}
\end{figure}

The task of apple harvesting requires high accuracy in the final approach, combined with a short path. 
To calculate the accuracy in the final approach, the first step was to define the final approach. 
The start of the final approach is indicated in Fig.~\ref{fig:task-example} using a purple line. In the end-frame, this point lies at Y=0.05 m. This point was chosen as it indicates the position from where collisions with the target can occur, which would result in failed grasps.
A close-up of the final approach is shown in Fig.\ref{fig:final-approach-example}. Once the end-effector is closer than 5 cm along the Y-axis, the remaining section is the final approach of the trajectory. Within the final approach, the pose of each timestep is evaluated for being within the accuracy boundaries, indicated in orange. If the pose is within the boundaries for all dimensions, that pose is counted as accurate. The final approach accuracy is the number of accurate poses divided by the total number of poses in the final approach. 
As all demonstrations were successful, we used the median absolute deviations from the origin as the limits for the final approach phase. Staying below these limits indicates a successful final approach phase in the trajectory of an execution. These limits were 0.03 m along the X-axis, 0.10 m along the Z-axis, 0.07 rad along the roll axis, 0.07 rad along the pitch axis, and 0.07 along the yaw axis.
We used the fraction in the demonstrations, which was at 0.88, as the required accuracy.
For the requirement of a short path, we looked at the distance the manipulator travelled to perform the task. This was calculated using the Euclidean distance in meters and the quaternion distance in radians. These were calculated for all timesteps. This results in the following equation:

\begin{equation}\label{eq:distance}
    d = \sum_{t=1}^{T}\left( \vert p_t - p_{t+1}\rvert + 
    2\arccos \left( \vert q_t \cdot q_{t+1} \rvert \right) \right)
\end{equation}

\noindent With \(p\) being the position vector and \(q\) being the orientation, represented as a quaternion.

Based on this analysis, we compared the optimal settings of both methods on key elements of the task. We compared the methods on final approach accuracy, Euclidean and rotation distance as included in Equation~\ref{eq:distance}.

\subsection{Set-up of apple grasping experiment}
\begin{figure}[t]
    \centering
    \includegraphics[width=0.45\columnwidth]{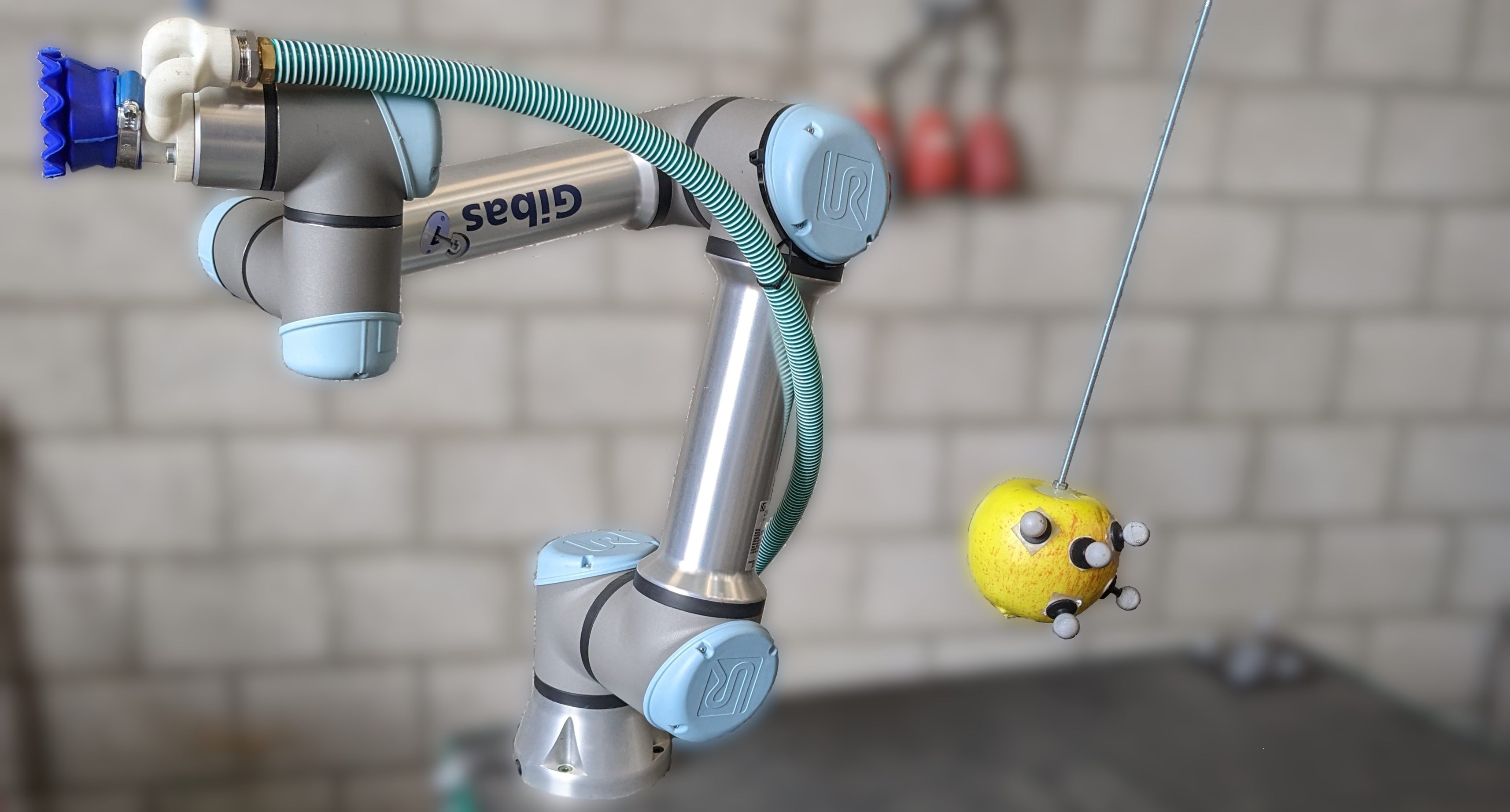}
    \caption{Initial state of the apple grasping set-up}
    \label{fig:apple-set-up}
\end{figure}

Next, we tested the performance of the method in a real-world environment while doing a harvesting task, specifically apple grasping using a suction cup gripper, using the same demonstration set. Fig.~\ref{fig:apple-set-up} shows the initial state of the apple grasping set-up. On the left, the UR5e manipulator is shown. The end-effector is a suction cup. On the right, the apple is shown. We used a fake apple for the experiments, with OptiTrack~\cite{RN39} markers to track its pose. The apple was attached to a rod, oscillating in a pendulum motion, along the X-axis of the end-frame.  Before each run, the apple was positioned at one of the extremes of the pendulum motion. At the start of each run, the apple was released, resulting in a dampened oscillating motion. Compared to the simulation experiment, the oscillation of the apple corresponds to the combination of a medium position oscillation and a medium orientation oscillation. 

The goal of this experiment was to grasp the apple successfully. In addition to the aforementioned metrics of final accuracy and distance travelled, we evaluated the time until successful grasping. A successful grasp happened when the air pressure in the suction cup dropped below 0.9 bar. If this happened before the trajectory was completed, the time the pressure dropped was used as the time until grasping. 
Otherwise, the time to execute the trajectory was stored.  

\section{Results} \label{sec:results}

\subsection{Results of simulation experiment}

\begin{figure}[t]
    \centering
    \subfloat[InfLQR]{
        \includegraphics[width=0.32\columnwidth]{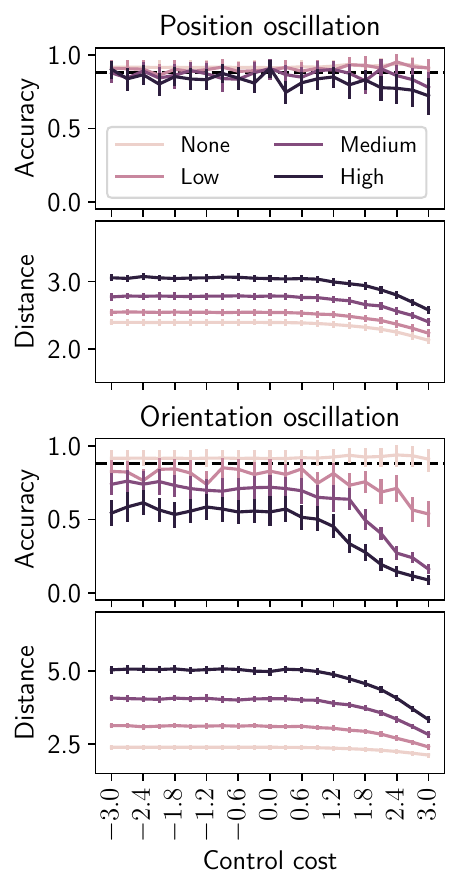}
        \label{fig:res-inf-lqr}
    }
    \subfloat[Single LQR]{
        \includegraphics[width=0.32\columnwidth]{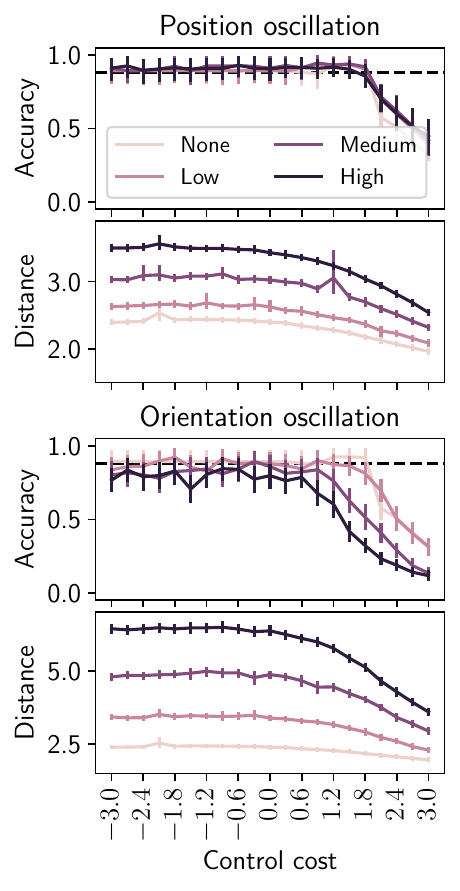}
        \label{fig:res-single-lqr}
    }
    \subfloat[Dual LQR]{
        \includegraphics[width=0.32\columnwidth]{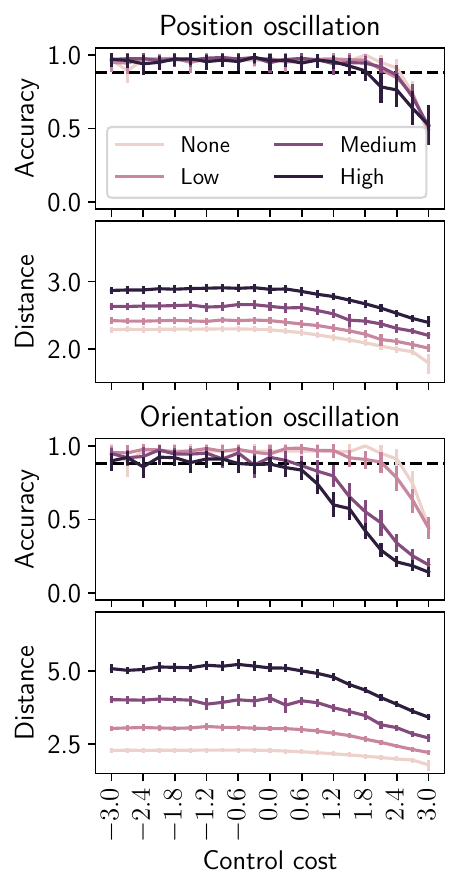}
        \label{fig:res-dual-lqr}
    }
    \addtolength{\abovecaptionskip}{0.05in}
    \caption{Results of the simulation testing of both methods. Darker lines indicate higher oscillation levels. The levels are described in section~\ref{sec:sim-exp}. The figures on the left show the effect of an oscillation of the target's position, and the figures on the right show the effect of an oscillation of the target's orientation. At each tested setting, a vertical bar indicates the 95\% confidence interval. The black dashed line indicates the required accuracy, set at 0.88.}
    \label{fig:res-lqr}
\end{figure}

Fig.~\ref{fig:res-lqr} shows the performance of InfLQR, a single LQR and DualLQR in terms of accuracy and distance travelled as a function of the control cost \(\rho\) used. 

For all controllers, increasing oscillations increased the distance travelled, and increasing the control cost reduced the distance travelled. The single LQR had the largest distance increase, and InfLQR and DualLQR were quite similar. 

For InfLQR, with increasing position oscillations, the accuracy dropped slightly. With high oscillations, only two controller settings reached the required accuracy. A slight effect of control cost was observed, with higher control costs resulting in lower accuracy. 
The InfLQR was unable to reach the required accuracy for any control cost.
For a single LQR, high values of the control cost resulted in reduced accuracy in the final approach. Without oscillations, the control cost needed to be \(1.8\) or lower. Increased position oscillations had a minimal influence. Only with high oscillations the control cost needed to be \(1.5\) or lower.
Introducing oscillations around the orientation axes resulted in similar behaviour, but the effect on accuracy and distance was greater. At high oscillations, the required accuracy could not be reached. 
For DualLQR, the effect of the oscillations on the accuracy was the smallest.
Without oscillations, high control costs resulted in lower accuracy. To reach the required accuracy, the control cost needed to be \(2.4\) or lower. 
Introducing oscillations along the position of the target had a minimal effect. The value of control cost from which the required accuracy was reached dropped. For low and medium oscillations, this was \(2.1\). For high oscillations, this was \(1.8\).
Introducing oscillations along the orientation of the target resulted in a similar behaviour, but also affected the accuracy of low control costs. 
For low oscillations, the requirement was met at a control cost of \(2.1\) or lower. 
For medium oscillations, the controller reached the requirement from a control cost of \(0.3\) or lower, except for \(-0.3\). For high oscillations, the controller reached the requirement from a control cost of \(0.0\) or lower, except for \(-0.3\). 

To compare the methods, we selected the highest control cost that reached the required accuracy. 
For the InfLQR, we selected a control cost of \(0.0\), which was the highest control cost that reached the required accuracy under high position oscillations. 
For the single LQR, we selected a control cost of \(-0.3\), which was the control cost that reach the required accuracy under medium orientation oscillations. 
For the DualLQR, we selected a control cost of \(0.0\) as well, this was the highest control cost that reached the required accuracy under high orientation oscillations. 
Table~\ref{tab:best_models} shows the performance of both methods for all oscillations. DualLQR resulted in higher accuracy for all oscillations, increasing on average 18\% compared to InfLQR and 7.0\% compared to a single LQR. For the Euclidean distance, similar results were observed, with an average decrease of 3.9\% compared to InfLQR and 26\% compared to a single LQR. For rotated distance, the DualLQR decreased the distance with an average of 2.8\% compared to InfLQR and 6.8\% compared to a single LQR. The InfLQR had lower rotated distance for medium and high orientation oscillations, but this also resulted in much lower final approach accuracy.

\begin{table}[t]
    \centering
    \caption{Performance in simulation of best models for each oscillation level \& method. Each value shows the average and standard deviation. Bold values indicate the best method. Superscripts indicate grouping using Tukey's HSD test.}
    \label{tab:best_models}
    \begin{tabular*}{\linewidth}{@{\extracolsep{\fill}} llrrr}
        \toprule
         Oscillation & Method & Accuracy & Translation & Rotation \\
        \midrule
        \multirow[t]{3}{*}{None} & DualLQR & \textbf{0.96\(\pm\)0.11\textsuperscript{a}} & \textbf{0.59\(\pm\)0.039\textsuperscript{a}} & \textbf{1.69\(\pm\)0.073\textsuperscript{a}} \\
         & InfLQR & \textbf{0.91\(\pm\)0.15\textsuperscript{ab}} & 0.62\(\pm\)0.033\textsuperscript{b} & 1.77\(\pm\)0.076\textsuperscript{c} \\
         & Single LQR & 0.89\(\pm\)0.23\textsuperscript{b} & 0.68\(\pm\)0.048\textsuperscript{c} & 1.73\(\pm\)0.060\textsuperscript{b} \\
        \midrule
        \multirow[t]{3}{*}{Low, Orientation} & DualLQR & \textbf{0.95\(\pm\)0.16\textsuperscript{a}} & \textbf{0.61\(\pm\)0.041\textsuperscript{a}} & \textbf{2.42\(\pm\)0.119\textsuperscript{a}} \\
         & InfLQR & \textbf{0.83\(\pm\)0.23\textsuperscript{a}} & \textbf{0.63\(\pm\)0.033\textsuperscript{a}} & \textbf{2.47\(\pm\)0.114\textsuperscript{a}} \\ 
         & Single LQR & \textbf{0.89\(\pm\)0.21\textsuperscript{a}} & 0.77\(\pm\)0.138\textsuperscript{b} & 2.71\(\pm\)0.208\textsuperscript{b} \\
        \midrule
        \multirow[t]{3}{*}{Low, Position} & DualLQR & \textbf{0.98\(\pm\)0.10\textsuperscript{a}} & \textbf{0.73\(\pm\)0.043\textsuperscript{a}} & \textbf{1.69\(\pm\)0.072\textsuperscript{a}} \\
        & InfLQR & \textbf{0.90\(\pm\)0.15\textsuperscript{a}} & \textbf{0.76\(\pm\)0.039\textsuperscript{a}} & 1.78\(\pm\)0.075\textsuperscript{b} \\
        & Single LQR & \textbf{0.90\(\pm\)0.23\textsuperscript{a}} & 0.92\(\pm\)0.123\textsuperscript{b} & \textbf{1.74\(\pm\)0.118\textsuperscript{ab}} \\
        \midrule
        \multirow[t]{3}{*}{Medium, Orientation} & DualLQR & \textbf{0.92\(\pm\)0.15\textsuperscript{a}} & \textbf{0.66\(\pm\)0.081\textsuperscript{a}} & \textbf{3.41\(\pm\)0.261\textsuperscript{a}} \\
        & InfLQR & 0.72\(\pm\)0.20\textsuperscript{b} & \textbf{0.66\(\pm\)0.039\textsuperscript{a}} & \textbf{3.39\(\pm\)0.192\textsuperscript{a}} \\
        & Single LQR & \textbf{0.89\(\pm\)0.20\textsuperscript{a}} & 0.91\(\pm\)0.088\textsuperscript{b} & 3.85\(\pm\)0.487\textsuperscript{b} \\
        \midrule
        \multirow[t]{3}{*}{Medium, Position} & DualLQR & \textbf{0.95\(\pm\)0.16\textsuperscript{a}} & \textbf{0.94\(\pm\)0.102\textsuperscript{a}} & \textbf{1.69\(\pm\)0.090\textsuperscript{a}} \\
        & InfLQR & \textbf{0.91\(\pm\)0.19\textsuperscript{a}} & 1.00\(\pm\)0.046\textsuperscript{b} & 1.79\(\pm\)0.080\textsuperscript{b} \\
        & Single LQR & \textbf{0.92\(\pm\)0.20\textsuperscript{a}} & 1.31\(\pm\)0.088\textsuperscript{c} & \textbf{1.73\(\pm\)0.060\textsuperscript{a}} \\
        \midrule
        \multirow[t]{3}{*}{High, Orientation} & DualLQR & \textbf{0.88\(\pm\)0.13\textsuperscript{a}} & \textbf{0.65\(\pm\)0.044\textsuperscript{a}} & \textbf{4.44\(\pm\)0.279\textsuperscript{a}} \\
        & InfLQR & 0.55\(\pm\)0.25\textsuperscript{b} & \textbf{0.69\(\pm\)0.053\textsuperscript{a}} & \textbf{4.28\(\pm\)0.271\textsuperscript{a}} \\
        & Single LQR & \textbf{0.77\(\pm\)0.24\textsuperscript{a}} & 1.05\(\pm\)0.101\textsuperscript{b} & 5.28\(\pm\)0.394\textsuperscript{b} \\
        \midrule
        \multirow[t]{3}{*}{High, Position} & DualLQR & \textbf{0.96\(\pm\)0.14\textsuperscript{a}} & \textbf{1.20\(\pm\)0.078\textsuperscript{a}} & \textbf{1.69\(\pm\)0.094\textsuperscript{a}} \\
        & InfLQR & \textbf{0.91\(\pm\)0.14\textsuperscript{a}} & \textbf{1.23\(\pm\)0.054\textsuperscript{a}} & 1.81\(\pm\)0.078\textsuperscript{b} \\
        & Single LQR & \textbf{0.91\(\pm\)0.21\textsuperscript{a}} & 1.75\(\pm\)0.084\textsuperscript{b} & \textbf{1.72\(\pm\)0.064\textsuperscript{a}} \\
        \bottomrule
        \bottomrule
    \end{tabular*}
\end{table}

\subsection{Results of apple grasping experiment}
\begin{figure}[t]
    \centering
    \includegraphics[width=\columnwidth]{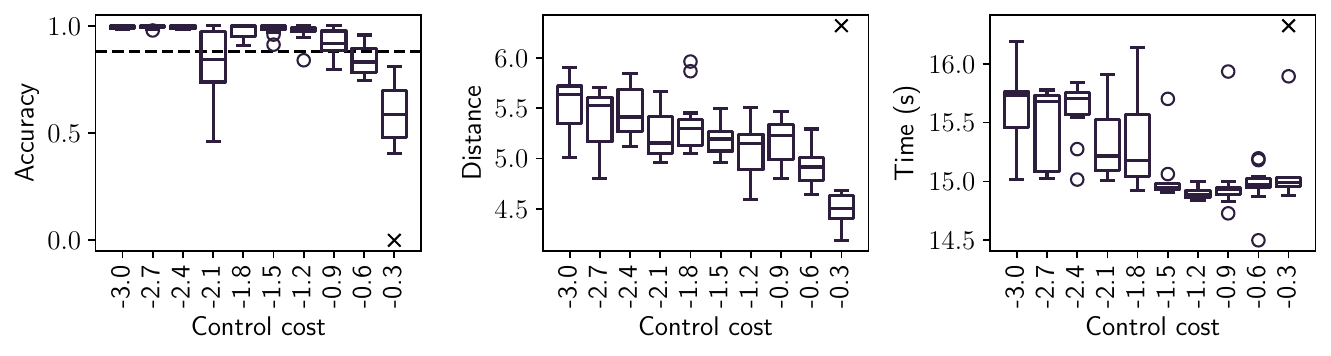}
    \addtolength{\abovecaptionskip}{-0.25in}
    \caption{Boxplots of results of testing the Dual LQR on apple grasping. X indicates a failed attempt.}
    \label{fig:apple-dual-lqr}
\end{figure}

When testing the performance of DualLQR on apple grasping, we found a very high success rate. Only a single grasp attempt out of 110 was unsuccessful, at a control cost of \(-0.3\), shown with a cross in Fig.~\ref{fig:apple-dual-lqr}. For this control cost, the final approach accuracy was also insufficient, as shown in Fig.~\ref{fig:apple-dual-lqr}. From -0.9, the required accuracy was met, except for -2.1. At this control cost, the yaw of the apple was around \(0.3\) less than the other settings.

In this experiment, the control cost needed to be lower to achieve the required accuracy, as shown in the left plot in Fig.~\ref{fig:apple-dual-lqr}. However, similar accuracy was achieved with low control costs. Furthermore, the distance travelled by the manipulator was nearly doubled compared to the simulation, as shown in the middle plot in Fig.~\ref{fig:apple-dual-lqr}. 
Lastly, the right plot of Fig.~\ref{fig:apple-dual-lqr} shows that the time until successful grasping was the lowest at a control cost of \(-1.2\). At a control cost of \(-1.8\) or lower, there was a lot of variation, while control costs above \(-1.2\) were slightly slower. 

\begin{figure}[t]
    \centering
    \subfloat[Control cost of \(-0.3\)]{\scalebox{0.45}{\includegraphics{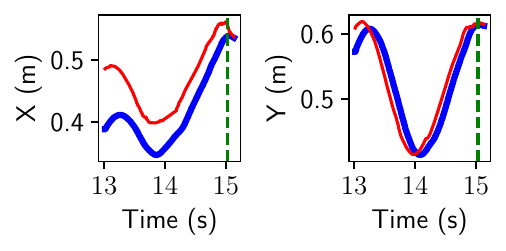}}\label{fig:approach-03}}
    \hfill
    \subfloat[Control cost of \(-1.2\)]{\scalebox{0.45}{\includegraphics{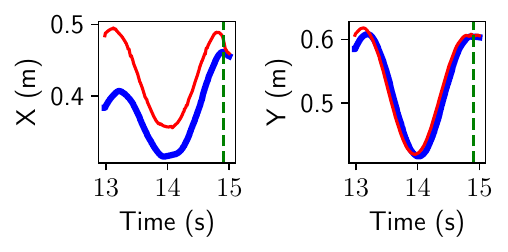}}\label{fig:approach-12}}
    \hfill
    \subfloat[Control cost of \(-3.0\)]{\scalebox{0.45}{\includegraphics{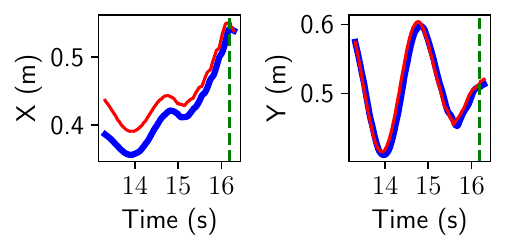}}\label{fig:approach-30}}
    \addtolength{\abovecaptionskip}{0.05in}
    \caption{Trajectory of final approach along X and Y. The red line indicates the pose of the target and the blue line indicates the pose of the robot. The green vertical line indicates the time of successful grasping. }
    \label{fig:apple-approach}
\end{figure}

Both effects were caused by the interaction between the suction cup and apple, shown in Fig.~\ref{fig:apple-approach}.
Fig.~\ref{fig:approach-03} shows the behaviour at a control cost of \(-0.3\). 
In this case, the apple was not tracked close enough for successful grasping. Instead, the side of the suction cup collided with the apple. This contact between the apple and the suction cup occurred close to the grasping. In the contact, the apple slipped over the suction cup, reducing speed and rotating slightly, which occurred just after 14 seconds, where the red line has sudden changes in the Y and X plot. 
Fig.~\ref{fig:approach-30} shows the behaviour at a control cost of \(-3.0\). 
In this case, the apple was tracked accurately. If the apple was not immediately grasped at first contact, the manipulator would push the apple away and react quickly, pushing it further away. This contact between the apple and the suction cup occurred multiple times close to the grasping. In this contact, the apple bounced off the flexible suction cup until the contact was good enough to create a vacuum. This can be seen in Fig.~\ref{fig:approach-30}, where there are multiple sudden changes in the X plot, as the fruit is moved further away. 
Fig.~\ref{fig:approach-12} shows the behaviour at a control cost of \(-1.2\). With control cost \(-1.2\), these two effects were balanced optimally, resulting in the fastest grasping. No perceivable contact occurred between the apple and the suction cup until the grasp. This can be seen in the smooth trajectory of the target in Fig.~\ref{fig:approach-12}.

\section{Discussion} \label{sec:discussion}
DualLQR increased the final approach accuracy and reduced the distance travelled compared to both InfLQR and using a single LQR. 
Compared to the InfLQR, this was due to the use of the finite horizon LQR and reducing the computations in the real time loop, allowing for faster responses to changes in the target pose. In addition, the difference in required precision between the reference frames was used twice. First, in the fitting of each LQR, where higher precision resulted in larger control outputs. Next, it was used again when combining the outputs. By using the difference in required precision twice, more emphasis was placed on the frame with higher required precision. 
Compared to the single LQR, the DualLQR also increased the accuracy, while the opposite was expected. The additional reference frame of the DualLQR will influence the movement even during the final approach. Most noticeably, the DualLQR had less variation. This indicates that the additional reference frame dampened the control output, resulting in the more accurate approach. 

In order to evaluate the performance of the methods, we used the final approach accuracy as a metric. Related work primarily used the accuracy of the final pose~\cite{RN198,RN179}. However, this does not take the approach into account, where collisions can result in additional movement of a target or damage to a target.
However, the final-approach-accuracy metric also leaves room for improvement. In the apple grasping experiment, the DualLQR was able to grasp apples successfully despite not meeting the required accuracy. The metric can be adjusted to calculate deviation from a desired path. However, a desired path would need to be defined.

The experiments provided an empirical evaluation of the system robustness. There are systemic limitations to the system's robustness because of workspace and acceleration limitations of the manipulator. The acceleration limits of the fruit are likely lower, as the fruit would become detached otherwise. The selection of target fruits needs to consider where in the workspace the fruit is hanging to ensure the fruit will stay within the workspace even under oscillations. 

In the simulation experiment, the oscillation did not decay, while this would have been more realistic and in line with the apple grasping experiment. 

In the apple grasping experiment, the OptiTrack system and manipulator added roughly 45 ms of latency to the control loop between a movement of the target in the world and an action of the manipulator. Combined with the dynamics required to move the manipulator, this resulted in a reduced accuracy compared to the simulation experiment. Using a camera and processing to obtain the pose of the fruit will add even more time. This latency could be addressed by using a predictor to estimate the future pose of the fruit, e.g. by using Model Predictive Control~\cite{RN208}. This can improve the performance of the proposed method in the apple grasping experiment, where the apple was oscillating consistently. However, in the case of a real apple tree, this prediction will be more challenging, due to variable flexibility in branches, wind, and other operations in the same tree. Additionally, it would increase the latency even more, which reduces its applicability. In our experiments, we found that lower control costs result in similar performance between the simulation and real environment. In a system with increased latency, even lower control costs might be required. 

Through the OptiTrack system, we obtained a continuous and precise pose of the apple. In a real scenario, a vision system will be less reliable. 
When far away, the influence of incorrect pose estimation will be minimized. As the robot approaches the fruit, the pose estimation is expected to improve, and DualLQR can adjust for these improvements.
However, when very close to the fruit, the robot can block the view of the fruit. Therefore, further sensorization of the system might be needed to obtain the pose of the apple when close to the fruit. 

The grasp of the apples was detected based on the pressure value in the suction cup. While no false positives occurred during our experiments, additional sensors can be implemented to provide a second measurement of fruit pressure. These could use a FT-sensor to detect the weight of the fruit in the suction cup, or measure the deformation of the suction cup. Alternatively, as long as the false positive does not result in damage, the pick can be attempted again. 

In the apple grasping experiment, a reduced accuracy did not yet lead to reduced grasping success. 
The suction cup was continuously sucking and showed a tolerance for misalignment of around 1 cm. If the apple was within the tolerance at some point during the action, the apple would be grasped. This effect increased the success rate. 
However, contact also occurred with greater misalignment, which caused the apple to bounce off. At low control costs, repeated bouncing could occur due to the fast movement of the manipulator after the initial bounce.
While the contact itself is not expected to cause damage to real apples due to the use of soft silicon gripper, the movement caused by the interaction could cause damage to the branch or peduncle. 
In addition, higher accuracy would be required if an encompassing gripper was used.

For the DualLQR, we used two reference frames. The method can be extended to include any number of reference frames. This can even be extended to include for example obstacle avoidance strategies, if an algorithm provides the importance of the obstacle avoidance through a covariance matrix. 

To facilitate the reproducibility of this work, we made the code to use the proposed method available, including the set-up for the simulation and apple grasping experiment. The datasets and trained models are available upon request. Most of the set-up consisted of off-the-shelf components. The end-effector was custom-made, for which the 3D models can be made available upon request.
As the performance of the model depends on the demonstrations provided, training the models with a new set of demonstrations will influence the achievable performance. 

\section{Conclusion} \label{sec:conclusion}
In this work, we presented DualLQR, a novel Learning from Demonstration method aimed at approaching oscillating targets while only tracking the oscillation near the target. 
Through extensive simulation testing, it was found that InfLQR barely meets the required final accuracy without oscillations and drops below the required accuracy with an oscillating target. On the other hand, a single LQR reached the required accuracy, but required a large increase in the distance travelled to reach the target.
DualLQR was found  to meet the required final accuracy even with high oscillations, while travelling the least distance.
Compared to InfLQR, DualLQR increased the accuracy by 18\%, reduced the translated distance by 3.7\% and the rotated distance by 2.8\%. 
Compared to a single LQR, DualLQR increased the accuracy by 7.0\%, reduced the translated distance by 26\% and the rotated distance by 6.8\%.

Further testing in a simplified apple grasping experiment showed that DualLQR can successfully grasp oscillating apples, with a success rate of 99\%. The optimal control cost was found to be \(-1.2\). Here, grasping the apple was done in 14.9 seconds on average, compared to 15.6 seconds for the slowest control cost setting. In addition, the fruit was not touched before grasping.

\begin{credits}
\subsubsection{\ackname} This work was partially funded by the Dutch Ministry of LVVN (Agriculture, Fisheries, Food Security, and Nature), under the project code KB-38-001-005, and by the Netherlands Organisation for Scientific Research (NWO grant 17626, project Synergia).

\subsubsection{\discintname} The authors have no competing interests to declare that are relevant to the content of this article.

\subsubsection{Electronic supplementary material} A supporting video can be found at \url{https://youtu.be/2bY84MN53tA}

\end{credits}

\bibliography{paper}

\end{document}